\documentclass[conference,9pt]{IEEEtran} 
\IEEEoverridecommandlockouts
\usepackage{cite}
\usepackage{amsmath,amssymb,amsfonts}
\usepackage{algorithmic}
\usepackage{graphicx}
\usepackage{textcomp}
\usepackage{xcolor}
\usepackage{subcaption}
\usepackage{pgfplots}
\usepackage{tikz}
\pgfplotsset{compat=newest}
\usetikzlibrary{plotmarks}
\usetikzlibrary{arrows.meta,positioning}
\usetikzlibrary{shapes,arrows, positioning, automata}
\usepackage[hidelinks]{hyperref}

\tikzset{cross/.style={cross out, draw=black, minimum size=2*(#1-\pgflinewidth), inner sep=0pt, outer sep=0pt},cross/.default={4pt}}

\usepgfplotslibrary{patchplots}
\usepackage{grffile}

\usepackage{multirow}
\def\BibTeX{{\rm B\kern-.05em{\sc i\kern-.025em b}\kern-.08em
    T\kern-.1667em\lower.7ex\hbox{E}\kern-.125emX}}

\begin{document}

\title{Increasing the Accuracy of a Neural Network Using Frequency Selective Mesh-to-Grid Resampling} 

\author{\IEEEauthorblockN{Andreas Spruck,
Viktoria Heimann and
Andr\'e Kaup}
\IEEEauthorblockA{Multimedia Communications and Signal Processing\\
Friedrich-Alexander-Universität Erlangen-Nürnberg,\\
Cauerstr. 7, 91058 Erlangen, Germany \\ 
Email: \{andreas.spruck, viktoria.heimann, andre.kaup\}@fau.de
}}

\maketitle

\begin{abstract}
Neural networks are widely used for almost any task of recognizing image content. Even though much effort has been put into investigating efficient network architectures, optimizers, and training strategies, the influence of image interpolation on the performance of neural networks is not well studied. Furthermore, research has shown that neural networks are often sensitive to minor changes in the input image leading to drastic drops of their performance. Therefore, we propose the use of keypoint agnostic frequency selective mesh-to-grid resampling (FSMR) for the processing of input data for neural networks in this paper. This model-based interpolation method already showed that it is capable of outperforming common interpolation methods in terms of PSNR. Using an extensive experimental evaluation we show that depending on the network architecture and classification task the application of FSMR during training aids the learning process. Furthermore, we show that the usage of FSMR in the application phase is beneficial. 
The classification accuracy can be increased by up to 4.31 percentage points for ResNet50 and the Oxflower17 dataset. 
\end{abstract}


\section{Introduction}
Neural networks are a common and widely used method to process image content nowadays. The strength of neural networks is to extract information provided by the image content. This is a task that may be easily solved by humans but is very hard to describe analytically. With the usage of neural networks for the extraction of information provided by the content of an image, the whole strategy of solving classification problems changed. Neural networks get rid of the necessity to analytically describe the task, but learn to extract the desired information from an image by training. Many images with accompanying labels are shown to the neural network during the training process. Due to the combination of image and label, the network detects the features describing the desired information by itself. \par
During the last years, much effort was put into finding efficient, robust, and accurate architectures for neural networks \cite{Krizhevsky2012,Simonyan2015,Szegedy2016,He2016a,Szegedy2017,Huang2017,Sandler2018,Tan2019}. Many training strategies evolved with the aim to obtain best possible results very fast \cite{Wong2016,Mikolajczyk2018}. Furthermore, increasingly large datasets for all kinds of application cases were proposed \cite{Nilsback2006,Parkhi2012,FeiFei2004}. These developments drove the recent success of neural networks. An area of research that has not been examined up to this extent is the influence of image interpolation methods on the final results. \par
Image interpolation is a processing step occurring in almost every training and application pipeline of neural networks as most networks require an input image of predefined size. 
Thus, in order to combine every dataset with every network, affine transforms such as resizing have to be applied to the original input images. Depending on the use case, further processing steps are performed on the original input image, e.~g., lens correction or rectification. Therefore, interpolation is required. \par
\begin{figure}
\centering
\resizebox{0.23\textwidth}{!}{
\begin{tikzpicture}
\draw[fill=black](5,5)circle(2pt);
\draw[fill=black](5,1)circle(2pt);
\draw[fill=black](5,2)circle(2pt);
\draw[fill=black](5,3)circle(2pt);	
\draw[fill=black](5,4)circle(2pt);	
\draw[fill=black](1,5)circle(2pt);
\draw[fill=black](1,1)circle(2pt);	
\draw[fill=black](1,2)circle(2pt);
\draw[fill=black](1,3)circle(2pt);	
\draw[fill=black](1,4)circle(2pt);
\draw[fill=black](2,5)circle(2pt);
\draw[fill=black](2,1)circle(2pt);
\draw[fill=black](2,2)circle(2pt);
\draw[fill=black](2,3)circle(2pt);	
\draw[fill=black](2,4)circle(2pt);	
\draw[fill=black](3,5)circle(2pt);
\draw[fill=black](3,1)circle(2pt);	
\draw[fill=black](3,2)circle(2pt);
\draw[fill=black](3,3)circle(2pt);	
\draw[fill=black](3,4)circle(2pt);
\draw[fill=black](4,5)circle(2pt);
\draw[fill=black](4,1)circle(2pt);	
\draw[fill=black](4,2)circle(2pt);
\draw[fill=black](4,3)circle(2pt);	
\draw[fill=black](4,4)circle(2pt);	

\draw[step=1.0,black,thin] (1,1) grid (5,5);

\draw[color=red](	 0.2679  ,  2.2679 )circle(2pt);
\draw[color=red](    1.1340  ,  1.7679 )circle(2pt);
\draw[color=red](    2.0000  ,  1.2679 )circle(2pt);
\draw[color=red](    2.8660  ,  0.7679 )circle(2pt);
\draw[color=red](    3.7321  ,  0.2679 )circle(2pt);
\draw[color=red](    0.7679  ,  3.1340 )circle(2pt);
\draw[color=red](    1.6340  ,  2.6340 )circle(2pt);
\draw[color=red](    2.5000  ,  2.1340 )circle(2pt);
\draw[color=red](    3.3660  ,  1.6340 )circle(2pt);
\draw[color=red](    4.2321  ,  1.1340 )circle(2pt);
\draw[color=red](    1.2679  ,  4.0000 )circle(2pt);
\draw[color=red](    2.1340  ,  3.5000 )circle(2pt);
\draw[color=red](    3.0000  ,  3.0000 )circle(2pt);
\draw[color=red](    3.8660  ,  2.5000 )circle(2pt);
\draw[color=red](    4.7321  ,  2.0000 )circle(2pt);
\draw[color=red](    1.7679  ,  4.8660 )circle(2pt);
\draw[color=red](    2.6340  ,  4.3660 )circle(2pt);
\draw[color=red](    3.5000  ,  3.8660 )circle(2pt);
\draw[color=red](    4.3660  ,  3.3660 )circle(2pt);
\draw[color=red](    5.2321  ,  2.8660 )circle(2pt);
\draw[color=red](    2.2679  ,  5.7321 )circle(2pt);
\draw[color=red](    3.1340  ,  5.2321 )circle(2pt);
\draw[color=red](    4.0000  ,  4.7321 )circle(2pt);
\draw[color=red](    4.8660  ,  4.2321 )circle(2pt);
\draw[color=red](    5.7321  ,  3.7321 )circle(2pt);

\end{tikzpicture}
}
\caption{Example of the point distribution after a rotation by 30 degree. The black dotes denote the initial point locations on the grid structure. The red circles denote the point locations after the rotation. Interpolation methods are needed to resample the red circles lying on the mesh back onto a regular grid in order to display and store the resulting image.}
\label{fig:interp}
\vspace{-0.5cm}
\end{figure}
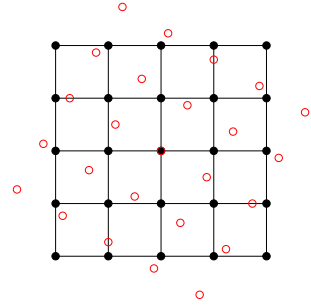%
Moreover, research has shown that training strategies such as data augmentation increase the performance of the network. Common data augmentation techniques are rotating, zooming or cropping of the input images. Figure~\ref{fig:interp} exemplary shows the distribution of points after rotating an image by thirty degrees. The initial pixel locations are denoted by black dots. The red circles indicate the resulting pixel locations after the rotation. The pixels after rotation have to be resampled onto the regular grid structure again in order to save and further process the image. For this task image interpolation is a crucial step. %
In the context of neural networks, common techniques are bilinear and bicubic interpolation. These interpolation methods are well known and fast, but not optimal in terms of the resulting image quality. With keypoint agnostic frequency selective mesh-to-grid resampling \cite{Heimann2020} an interpolation method was recently proposed that is capable of clearly outperforming common interpolation algorithms such as bilinear or bicubic interpolation for the desired affine transforms. Research has shown that neural networks are quite sensitive to small changes within an image \cite{MoosaviDezfooli2016}. Due to this behavior, even minor changes can lead to misclassifications \cite{Joshi2019}. Therefore, we investigate the influence of the used image interpolation technique on the overall result of a neural network classifier in this paper. \par
In the next section, we give an overview over the keypoint agnostic frequency selective mesh-to-grid resampling algorithm. 
In Section~\ref{sec:ex_setup}, we present the setup used for the experiments conducted. In Section~\ref{sec:evaluation}, we evaluate and discuss the results from our experiments, before the paper is concluded in Section~\ref{sec:conclusion}.%
%
%
%
\section{Frequency-Selective Mesh-to-Grid Resampling}
\label{sec:fsmr}
In the upcoming section, we will briefly introduce the keypoint agnostic frequency-selective mesh-to-grid resampling (FSMR). The core idea of FSMR is to block-wise generate a model of an image by superposition of weighted basis functions. The idea of using a frequency model for image interpolation has been extensively researched. This technique was able to show outstanding results for image reconstruction \cite{Genser2019} as well as for video reconstruction \cite{Spruck2019}. While these methods are restricted to data points lying on regular grid positions, FSMR gets rid of this constraint and is capable of generating a frequency model from arbitrarily distributed data points \cite{Heimann2020}. Therefore, FSMR iteratively generates a model
\begin{equation}
g^{(\nu)}[m,n] = g^{(\nu-1)}[m,n] + \hat{c}_{u,v} \varphi_{u,v}[m,n]
\end{equation}
with $\nu$ indicating the current iteration, $m,n$ being the pixel coordinates and $\hat{c}_{u,v}$ denoting the estimated expansion coefficient of the selected basis function $\varphi_{u,v}[m,n]$. The aim of this model is to match the original signal 
\begin{equation}
f[m,n] = \sum_{k,l \in \mathcal{K}} {c}_{k,l} \varphi_{k,l}[m,n]
\end{equation}
as good as possible. Here, $\mathcal{K}$ denotes the set of available basis functions and ${c}_{k,l}$ is defined as expansion coefficient. As bases functions from orthogonal transformations as in our case the discrete cosine transform are incorporated, the expansion coefficients can be understood as transformation coefficients. The generated model $g[m,n]$ should meet the original signal $f[m,n]$ as close as possible. Therefore, in every iteration $\nu$ the basis function is selected that maximizes the reduction of the residual energy $\Delta E_{k,l}^{(\nu)}$, following 
\begin{equation}
(u,v) = \mathrm{argmax}_{k,l}\left(\Delta E_{k,l}^{(\nu)} w_f[k,l] \right).
\end{equation}
The frequency weighting $w_f[k,l]$ %
favors low frequencies to reflect the typical local spectral characteristics of natural images \cite{Seiler2015}. \par
\begin{figure}[t]
\centering
	\begin{subfigure}[b]{0.25\columnwidth}
         \centering
         \includegraphics[width=\textwidth]{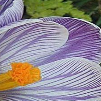}
         \caption{bilinear}
     \end{subfigure}
     \hfill
     \begin{subfigure}[b]{0.25\columnwidth}
         \centering
         \includegraphics[width=\textwidth]{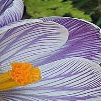}
         \caption{bicubic}
     \end{subfigure}
     \hfill
     \begin{subfigure}[b]{0.25\columnwidth}
         \centering
         \includegraphics[width=\textwidth]{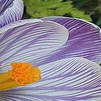}
         \caption{FSMR}
     \end{subfigure}
\caption{Comparison of the three used interpolation methods for a resizing operation shown on an excerpt from image 22 from class 'Crocus' of the \textit{Oxflower17} dataset. FSMR can reconstruct the structured lines on the petal best. Best viewed enlarged.}
\label{fig:comp_interp}
\vspace{-0.5cm}
\end{figure}

It could be shown that FSMR is capable of clearly outperforming common interpolation methods such as bilinear, bicubic and Lanczos interpolation in terms of PSNR and SSIM \cite{Heimann2020}. The experiments conducted in \cite{Heimann2020} explicitly evaluate image transformations as zoom and rotation which are of special interest in the scope of training deep neural networks, as will be shown in the next section. A visual example of the reconstruction quality can be seen in Figure~\ref{fig:comp_interp}. Here FSMR is capable of reconstructing the fine lines on the petal best. \par
In terms of processing time the FSMR approach is significantly slower than bilinear and bicubic interpolation, as for FSMR a non optimized Matlab implementation is used. Nevertheless, it was shown in recent publications that the processing time can be sped up drastically \cite{Heimann2020},\cite{Heimann2021}. Following \cite{Regensky2020} even a real time implementation of the algorithm is possible.
%
\section{Experimental Setup}
\label{sec:ex_setup}

%
%
In the following, we describe the setups of the experiments that we conducted. Besides the neural networks and datasets used during our experiments, we further present the pursued data augmentation strategies.
\subsection{Neural Networks}
For our experiments, five state-of-the-art networks were examined. During the selection of the networks special attention was payed to the overall size of the networks represented by the number of trainable parameters. In order to obtain results that may be generalized to other scenarios, networks from a large range of different sizes were chosen. The smallest network examined is from the widely used family of EfficientNets proposed in 2019 by Tan and Le \cite{Tan2019}. For our experiments we used the EfficientNetB0 which holds 5.3 millon parameters and is a small and efficient network. In 2017, Huang et. al. proposed the family of DenseNets as an efficient architecture that might overcome the vanishing gradient problem by connecting each layer to all preceding layers \cite{Huang2017}. For our evaluations, we used DenseNet121 holding 8 million parameters. The two largest networks examined are from the family of ResNets, that were proposed in 2016 by He et. al. \cite{He2016a}. ResNet50 represents a medium sized network with 25 million parameters in total, whereas ResNet152 represents a large network holding 60 million parameters. We use the implementations of the networks from \cite{He2016b} for our experiments. 
\subsection{Datasets}
We trained and evaluated the networks on three different datasets \textit{Oxflower17} \cite{Nilsback2006}, \textit{Oxford-IIIT-Pet} \cite{Parkhi2012}, and \textit{Caltech101} \cite{FeiFei2004}. The \textit{Oxflower17} dataset contains images of flowers from 17 different classes. The images contain variations in perspective, illumination, and aspect ratio even within the same class. The similar appearance of some of the flower species makes it a challenging classification task \cite{Nilsback2006}. Each class holds 80 images in total. During our experiments we used a 10\% test split resulting in 8 test images per class. 
The second dataset examined is the \textit{Oxford-IIIT-Pet} dataset. The dataset holds 37 different classes containing 12 breeds of cats and 25 breeds of dogs. Again, the images within the same class differ largely in terms of perspective, aspect ratio, illumination and background. The differences between the breeds can again be small. Some breeds differ only in minor details which makes the correct classification challenging \cite{Parkhi2012}. The number of images per class is not uniformly distributed. Most classes hold 200 samples, while there are two classes holding only 199 and 191 images. Again we choose a 10\% split for testing the network which corresponds to 20 images per class. %
The \textit{Caltech101} dataset is well knwon in terms of machine learning. It holds 102 classes in total containing a large variety of objects of our everyday life reaching from animals over vehicles to objects. The number of images is very unbalanced reaching from 40 to 800 images for some classes. Most classes hold around 50 samples. We use a uniformly distributed test set to keep the results easily interpretable. Again we used a 10\% test split which results in 5 test samples per class. The test and training split remains unchanged throughout all experiments.
\subsection{Data Augmentation}
Training the networks from scratch, we incorporated data augmentation. Therefore, we prepared the datasets beforehand following a uniform procedure that differs only in the used interpolation method. The training dataset was augmented in order to achieve better classification results. Furthermore, the aim is to evaluate the influence of the interpolation method used. The incorporated augmentation methods are zoom and rotation. During the augmentation procedure every training sample is rotated by an arbitrary angle between -45 and 45 degrees. The used rotation angles are the same for all three interpolation methods. After rotation, the images are resized to match the input resolution of the neural networks of 224 by 224 pixels. The procedure for the zoom augmentation is analog. The images are zoomed in or out by a random factor, followed by a resize operation to match the input resolution. Again, the randomly chosen zoom factors are the same for all interpolation methods. Additionally to the augmented versions of each training sample, a resized version of the original sample is included into the training dataset as well. We computed the resized images with the same interpolation method which was used for the augmentation. \par
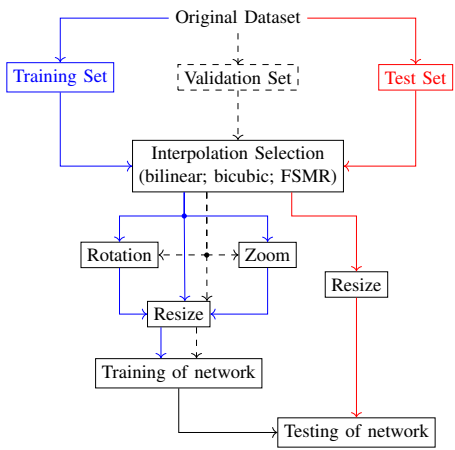
\begin{figure}
\centering
	\resizebox{0.7\columnwidth}{!}{
 	\begin{tikzpicture}

\node[align=center] at (5,7) (In) {Original Dataset};
\node[draw,blue,align=center] at (2,6) (Train) {Training Set};
\node[draw, dashed,align=center] at (5,6) (Val) {Validation Set};
\node[draw,red,align=center] at (8,6) (Test) {Test Set};
\node[draw,align=center] at (5,4.5) (Interp) {Interpolation Selection\\(bilinear; bicubic; FSMR)};

\node[draw,align=center] at (7,2.5) (Resize_test) {Resize};

\node[draw,align=center] at (3,3) (Rot) {Rotation};
\node[draw,align=center] at (5.5,3) (Zoom) {Zoom};
\node[draw,align=center] at (4,2) (Resize_train) {Resize};

\node[draw,align=center] at (4,1) (Train_net) {Training of network};
\node[draw,align=center] at (7,0) (Test_net) {Testing of network};


\coordinate [below=4mm of Interp.-154] (X1);
\filldraw[blue] (X1) circle (1pt);
\coordinate [below=10.6mm of Interp.-140] (X2); 
\filldraw[black] (X2) circle (1pt);
\coordinate [below=4mm of Interp.-26] (X3);

\coordinate [left=1mm of Resize_train.south] (Y1);
\coordinate [left=3mm of Resize_train.south] (Y2);
\coordinate [left=5mm of Resize_train.south] (Y3);

\coordinate [left=1mm of Train_net.north] (Y4);
\coordinate [left=3mm of Train_net.north] (Y5);
\coordinate [left=5mm of Train_net.north] (Y6);

\coordinate [below=5mm of Y2] (Y7);

\coordinate [right=1mm of Resize_train.south] (Z1);
\coordinate [right=3mm of Resize_train.south] (Z2);
\coordinate [right=5mm of Resize_train.south] (Z3);

\coordinate [right=1mm of Train_net.north] (Z4);
\coordinate [right=3mm of Train_net.north] (Z5);
\coordinate [right=5mm of Train_net.north] (Z6);

\coordinate [below=5mm of Z2] (Z7);

\draw [blue,->] (In) -| (Train.north);
\draw [blue,->] (Train) |- (Interp.west);
\draw [blue,->] (Interp.-154)-- (X1) -| (Rot.north);
\draw [blue,->] (Interp.-154) -- (X1) -| (Zoom.north);
\draw [blue,->] (Interp.-154) -- (X1) -- (Resize_train.65);
\draw [blue,->] (Zoom) |- (Resize_train.east);
\draw [blue,->] (Rot) |- (Resize_train.west);

\draw [blue,->] (Y2) -- (Y5);

\draw [dashed, ->] (In) -- (Val);
\draw [dashed, ->] (Val) -- (Interp);
\draw [dashed, ->] (Interp.-140) -- (X2) |- (Rot.east);
\draw [dashed, ->] (Interp.-140) -- (X2) |- (Zoom.west);
\draw [dashed, ->] (Interp.-140) -- (X2) -- (Resize_train.25); 


\draw [dashed, ->] (Z2) -- (Z5);

\draw [red,->] (In) -| (Test.north);
\draw [red,->] (Test) |- (Interp.east);
\draw [red,->] (Interp.-26) |- (X3) -| (Resize_test);
\draw [red,->] (Resize_test) -- (Test_net);

\draw [->] (Train_net) |- (Test_net);

\end{tikzpicture}
 	}
	\caption{Flow graph demonstrating the workflow during the data augmentation process. The workflow is the same for all datasets and is performed identically for each interpolation. The blue and dashed paths show the preparation pipeline for the training and validation dataset respectively. The red path denotes the processing pipeline for the test dataset.}
	\label{fig:flow_graph}
	\vspace{-0.5cm}	
\end{figure}

The samples of the test dataset are not augmented. Here, only a resizing to 224 by 224 pixels is performed in order to match the size of the input layer. The resizing is performed for all three interpolation methods separately, resulting in three independent test sets. Figure~\ref{fig:flow_graph} shows an overview over the process of preparing the different datasets. The blue path shows the preparation process conducted for the training dataset whereas the dashed path shows the processing pipeline for the validation dataset. The workflow for the training and validation samples is identical. With these samples the networks are trained and validated. The resulting trained networks are tested on the test dataset. The preparation of the test dataset is marked by the red path. 
\section{Evaluation}
\label{sec:evaluation}
\begin{table*}[ht]
	\centering
	\caption{Top-1 classification accuracy in \% for \textit{Oxflower17} for every combination of interpolation during test and training. The best result for each trained neural network is marked in bold font, same holds for the following tables.}
	\label{tab:results_flowers}
	\resizebox{0.98\textwidth}{!}{
		\begin{tabular}{|l|c|ccc|ccc|ccc|ccc|} \hline
			 \multicolumn{2}{|c}{} & \multicolumn{3}{|c}{EfficientNetB0} & \multicolumn{3}{|c}{DenseNet121} & \multicolumn{3}{|c}{ResNet50} & \multicolumn{3}{|c|}{ResNet152}\\ \cline{3-14}
			\multicolumn{2}{|c}{} & \multicolumn{3}{|c}{Test Interpolation} & \multicolumn{3}{|c}{Test Interpolation} & \multicolumn{3}{|c}{Test Interpolation} & \multicolumn{3}{|c|}{Test Interpolation} \\
			 \multicolumn{2}{|c|}{} & lin & cub & FSMR & lin & cub & FSMR & lin & cub & FSMR & lin & cub & FSMR \\ \hline
\multirow{3}{*}{\parbox{1.4cm}{Training \\ Interpolation}} & lin & 57.35 & 57.35 & \textbf{58.09} & \textbf{87.50} & {86.77} & {86.77} & {72.79} & 70.59 & \textbf{75.00} & \textbf{79.41} & \textbf{79.41} & \textbf{79.41} \\ \cline{2-14} 
			 			& cub & 52.94 & 52.94 & \textbf{55.15} & \textbf{83.82} & \textbf{83.82} & \textbf{83.82} & {75.74}& {75.74} & \textbf{76.47} & {76.47} & \textbf{77.21} & {75.74} \\ \cline{2-14} 
			 			& FSMR & {52.94} & 52.94 & \textbf{53.68} & {80.88} & 80.88 & \textbf{81.62} & {72.06} & {71.32} & \textbf{72.79} & {69.12} & {69.12} & \textbf{70.59} \\ \hline
		\end{tabular}
					}
\end{table*}
\begin{table*}[ht]
	\centering
	\caption{Top-1 classification accuracy in \% for \textit{Oxford-IIIT-Pet} for every combination of interpolation during test and training.}
	\label{tab:results_pets_37}
	\resizebox{0.98\textwidth}{!}{
		\begin{tabular}{|l|c|ccc|ccc|ccc|ccc|} \hline
			 \multicolumn{2}{|c}{} & \multicolumn{3}{|c}{EfficientNetB0} & \multicolumn{3}{|c}{DenseNet121} & \multicolumn{3}{|c}{ResNet50} & \multicolumn{3}{|c|}{ResNet152}\\ \cline{3-14}
			\multicolumn{2}{|c}{} & \multicolumn{3}{|c}{Test Interpolation} & \multicolumn{3}{|c}{Test Interpolation} & \multicolumn{3}{|c}{Test Interpolation} & \multicolumn{3}{|c|}{Test Interpolation} \\
			 \multicolumn{2}{|c|}{} & lin & cub & FSMR & lin & cub & FSMR & lin & cub & FSMR & lin & cub & FSMR \\ \hline
\multirow{3}{*}{\parbox{1.4cm}{Training \\ Interpolation}}&lin & \textbf{68.24} & 66.22 & 60.00 & \textbf{88.92} & 88.11 & 88.24 & 71.62 & 70.68 & \textbf{71.76} & \textbf{82.30} & 81.22 & \textbf{82.30} \\ \cline{2-14}
			 			&cub & \textbf{67.57} & 65.68 & 56.35 & 90.68 & {90.68} & \textbf{91.22} & 79.59 & \textbf{80.14} & 78.51 & \textbf{87.84} & 87.43 & {87.70} \\ \cline{2-14}
			 			&FSMR & 59.73 & 57.97 & \textbf{63.92} & 87.70 & 87.57 & \textbf{89.05} & {80.68} & 79.86 & \textbf{81.49} & 87.43 & 86.62 & \textbf{87.70} \\ \hline
		\end{tabular}
	}
\end{table*}
\begin{table*}[ht]
	\centering
	\caption{Top-1 classification accuracy in \% for \textit{Caltech101} for every combination of interpolation during test and training.} 
	\label{tab:caltech}
	\resizebox{0.98\textwidth}{!}{
		\begin{tabular}{|l|c|ccc|ccc|ccc|ccc|} \hline
			 \multicolumn{2}{|c}{} & \multicolumn{3}{|c}{EfficientNetB0} & \multicolumn{3}{|c}{DenseNet121} & \multicolumn{3}{|c}{ResNet50} & \multicolumn{3}{|c|}{ResNet152}\\ \cline{3-14}
			\multicolumn{2}{|c}{} & \multicolumn{3}{|c}{Test Interpolation} & \multicolumn{3}{|c}{Test Interpolation} & \multicolumn{3}{|c}{Test Interpolation} & \multicolumn{3}{|c|}{Test Interpolation} \\
			 \multicolumn{2}{|c|}{} & lin & cub & FSMR & lin & cub & FSMR & lin & cub & FSMR & lin & cub & FSMR \\ \hline
\multirow{3}{*}{\parbox{1.4cm}{Training \\ Interpolation}}&lin & \textbf{89.41} & 88.43 & 80.39 & \textbf{91.57} & 90.59 & 87.06 & \textbf{91.37} & 90.78 & 88.82 & \textbf{94.71} & 93.73 & 90.98\\ \cline{2-14}
			 			&cub & \textbf{89.61} & 89.22 & 80.78 & \textbf{94.90} & 94.51 & 92.16 & \textbf{91.37} & 90.59 & 79.22 & \textbf{89.02} & 87.65 & 84.31\\ \cline{2-14}
			 			&FSMR & 62.16 & 64.31 & \textbf{84.71} & 90.59 & 91.96 & \textbf{94.51} & 81.37 & 81.37 & \textbf{90.20} & 91.96 & 92.35 & \textbf{93.33}\\ \hline
		\end{tabular}
	}
\end{table*}
%
%
We evaluate the influence of the interpolation methods on the classification accuracy. 
We choose the commonly used bilinear (lin) and bicubic (cub) interpolation for the pre-processing of our datasets as described in Section~\ref{sec:ex_setup}. Additionally, we use FSMR that was introduced in Section~\ref{sec:fsmr}. During our extensive experiments we evaluated the performance of each neural network with each combination of interpolation during test and training for the three presented datasets \textit{Oxflower17} \cite{Nilsback2006}, \textit{Oxford-IIIT-Pet} \cite{Parkhi2012}, and \textit{Caltech101} \cite{FeiFei2004}. 
The number of test samples per class is equal for the entire dataset. \par 
Table~\ref{tab:results_flowers} shows the top-1 test accuracies, i.e. only correctly labeled images are accounted. The results are given in percent for all four neural networks and \textit{Oxflower17} dataset with the different training and test interpolations. Each row of the table corresponds to a certain interpolation used during training. The interpolation used for testing is given column-wise. 
The best test result is marked in bold font for each trained neural network. Examining the results given in Table~\ref{tab:results_flowers} it can be observed that using the proposed FSMR method on the test set is beneficial in nearly all examined scenarios. As the pre-processing of the test data using FSMR yields better results independent of the interpolation method used during the preceding training, it is a good choice for applications where the interpolation used during the initial training is not known. By this a gain of up to 4.31 percentage points in terms of classification accuracy can be achieved for ResNet50 trained using linear interpolation and tested on FSMR compared to the test set processed using bilinear interpolation. \par %
For the \textit{Oxford-IIIT-Pet} dataset we pursued an evaluation scheme distinguishing between all 37 different breeds of cats and dogs. The results for the classification on the different test sets are given in Table~\ref{tab:results_pets_37}. The structure of the table is the same as for the \textit{Oxflower17} dataset in Table~\ref{tab:results_flowers} and shows the classification accuracy for each network for every test and training combination of the three interpolation methods. Examining the results on the different test sets shown in Table~\ref{tab:results_pets_37}, we observe a similar behavior as for the \textit{Oxflower17} in Table~\ref{tab:results_flowers}. The larger networks DenseNet121, ResNet50, and ResNet152 benefit from a high quality interpolation such as FSMR during both training and test phase. This results in higher test accuracies as given in Table~\ref{tab:results_pets_37}. These networks also benefit from using a high quality interpolation such as FSMR during the pre-processing of the test data regardless of the interpolation used during training. On the other hand, EfficientNetB0 benefits from processing the dataset with bilinear interpolation. This can be explained by the network architecture. For the small EfficientNetB0 network, the input layer is a 3x3 convolution layer whereas for the remaining three networks it is a 7x7 convolution layer. As a consequence of that EfficientNetB0 has a smaller receptive field than the three larger networks. Due to the smaller receptive field the network is very sensitive towards local inconsistencies in the images. Therefore, EfficientNetB0 benefits from a smooth interpolation such as bilinear. \par
\begin{figure}
 	\centering
	\begin{subfigure}[b]{0.8\columnwidth}
         \centering
         \resizebox{\textwidth}{!}{
%
%
\begin{tikzpicture}
\pgfplotsset{every tick label/.append style={font=\Huge}}
\begin{axis}[%
width=10.293in,
height=3.652in,
at={(0in,0in)},
scale only axis,
unbounded coords=jump,
xmin=0.5,
xmax=17.5,
xtick={\empty},
ymin=10,
ymax=130,
xlabel=Classes,
ylabel = Standard Deviation,
label style={font=\Huge},
axis background/.style={fill=white}
]
\addplot [color=blue, forget plot]
  table[row sep=crcr]{%
1	31.0808915571398\\
1	81.3721458389253\\
};
\addplot [color=blue, forget plot]
  table[row sep=crcr]{%
2	39.1577104348395\\
2	83.6594979845987\\
};
\addplot [color=blue, forget plot]
  table[row sep=crcr]{%
3	43.6577031396532\\
3	105.262994180652\\
};
\addplot [color=blue, forget plot]
  table[row sep=crcr]{%
4	41.0956596379473\\
4	87.0593529102614\\
};
\addplot [color=blue, forget plot]
  table[row sep=crcr]{%
5	36.2092039821974\\
5	84.8093036578471\\
};
\addplot [color=blue, forget plot]
  table[row sep=crcr]{%
6	28.1864082099625\\
6	75.6247382321568\\
};
\addplot [color=blue, forget plot]
  table[row sep=crcr]{%
7	26.9029027874664\\
7	79.8676982299352\\
};
\addplot [color=blue, forget plot]
  table[row sep=crcr]{%
8	39.2349626097219\\
8	91.1688631221279\\
};
\addplot [color=blue, forget plot]
  table[row sep=crcr]{%
9	32.2164047325084\\
9	77.3330622392852\\
};
\addplot [color=blue, forget plot]
  table[row sep=crcr]{%
10	26.8046642811689\\
10	84.5951474741398\\
};
\addplot [color=blue, forget plot]
  table[row sep=crcr]{%
11	21.0035711808897\\
11	74.2816970020258\\
};
\addplot [color=blue, forget plot]
  table[row sep=crcr]{%
12	36.8224758888511\\
12	81.9041195859931\\
};
\addplot [color=blue, forget plot]
  table[row sep=crcr]{%
13	23.2733715556795\\
13	83.4332373160599\\
};
\addplot [color=blue, forget plot]
  table[row sep=crcr]{%
14	27.4458646369032\\
14	84.481970046359\\
};
\addplot [color=blue, forget plot]
  table[row sep=crcr]{%
15	29.1596530994208\\
15	74.0333154409922\\
};
\addplot [color=blue, forget plot]
  table[row sep=crcr]{%
16	23.3989397110321\\
16	73.195303222746\\
};
\addplot [color=blue, forget plot]
  table[row sep=crcr]{%
17	30.2791790756311\\
17	78.6971156746905\\
};
\addplot [color=blue, line width=4.0pt, forget plot]
  table[row sep=crcr]{%
1	47.9346514279201\\
1	64.9872509762128\\
};
\addplot [color=blue, line width=4.0pt, forget plot]
  table[row sep=crcr]{%
2	55.3434668804644\\
2	67.1314466495953\\
};
\addplot [color=blue, line width=4.0pt, forget plot]
  table[row sep=crcr]{%
3	62.7887326490041\\
3	81.9744342053312\\
};
\addplot [color=blue, line width=4.0pt, forget plot]
  table[row sep=crcr]{%
4	52.2680753655645\\
4	66.8881893862\\
};
\addplot [color=blue, line width=4.0pt, forget plot]
  table[row sep=crcr]{%
5	52.9683753384423\\
5	65.8971040510066\\
};
\addplot [color=blue, line width=4.0pt, forget plot]
  table[row sep=crcr]{%
6	45.4104619245263\\
6	58.5038478332937\\
};
\addplot [color=blue, line width=4.0pt, forget plot]
  table[row sep=crcr]{%
7	44.665164589972\\
7	65.4110913794234\\
};
\addplot [color=blue, line width=4.0pt, forget plot]
  table[row sep=crcr]{%
8	60.6873721914351\\
8	77.4342287179468\\
};
\addplot [color=blue, line width=4.0pt, forget plot]
  table[row sep=crcr]{%
9	47.7449612079145\\
9	60.6663806097384\\
};
\addplot [color=blue, line width=4.0pt, forget plot]
  table[row sep=crcr]{%
10	45.2445496502431\\
10	61.4800882774832\\
};
\addplot [color=blue, line width=4.0pt, forget plot]
  table[row sep=crcr]{%
11	37.0319797939804\\
11	53.8356293562719\\
};
\addplot [color=blue, line width=4.0pt, forget plot]
  table[row sep=crcr]{%
12	51.9957820591355\\
12	64.167063209605\\
};
\addplot [color=blue, line width=4.0pt, forget plot]
  table[row sep=crcr]{%
13	43.2272139351216\\
13	60.7657423126908\\
};
\addplot [color=blue, line width=4.0pt, forget plot]
  table[row sep=crcr]{%
14	45.4983447120713\\
14	61.3413160532537\\
};
\addplot [color=blue, line width=4.0pt, forget plot]
  table[row sep=crcr]{%
15	42.6786606541783\\
15	57.4765995502478\\
};
\addplot [color=blue, line width=4.0pt, forget plot]
  table[row sep=crcr]{%
16	40.9389339562241\\
16	56.0633366954477\\
};
\addplot [color=blue, line width=4.0pt, forget plot]
  table[row sep=crcr]{%
17	49.6154137275352\\
17	63.137400606185\\
};
\addplot [color=red, only marks, mark=*, mark options={solid, fill=red, draw=red, scale = 2}, forget plot]
  table[row sep=crcr]{%
1	55.3396460843456\\
};
\addplot [color=red, only marks, mark=*, mark options={solid, fill=red, draw=red, scale = 2}, forget plot]
  table[row sep=crcr]{%
2	61.2632653102473\\
};
\addplot [color=red, only marks, mark=*, mark options={solid, fill=red, draw=red, scale = 2}, forget plot]
  table[row sep=crcr]{%
3	74.6803838166018\\
};
\addplot [color=red, only marks, mark=*, mark options={solid, fill=red, draw=red, scale = 2}, forget plot]
  table[row sep=crcr]{%
4	60.3744127657698\\
};
\addplot [color=red, only marks, mark=*, mark options={solid, fill=red, draw=red, scale = 2}, forget plot]
  table[row sep=crcr]{%
5	59.2608964050586\\
};
\addplot [color=red, only marks, mark=*, mark options={solid, fill=red, draw=red, scale = 2}, forget plot]
  table[row sep=crcr]{%
6	50.6154294197305\\
};
\addplot [color=red, only marks, mark=*, mark options={solid, fill=red, draw=red, scale = 2}, forget plot]
  table[row sep=crcr]{%
7	54.1888245730239\\
};
\addplot [color=red, only marks, mark=*, mark options={solid, fill=red, draw=red, scale = 2}, forget plot]
  table[row sep=crcr]{%
8	66.6687824146424\\
};
\addplot [color=red, only marks, mark=*, mark options={solid, fill=red, draw=red, scale = 2}, forget plot]
  table[row sep=crcr]{%
9	53.9107833614023\\
};
\addplot [color=red, only marks, mark=*, mark options={solid, fill=red, draw=red, scale = 2}, forget plot]
  table[row sep=crcr]{%
10	50.6012539009079\\
};
\addplot [color=red, only marks, mark=*, mark options={solid, fill=red, draw=red, scale = 2}, forget plot]
  table[row sep=crcr]{%
11	46.1715732329023\\
};
\addplot [color=red, only marks, mark=*, mark options={solid, fill=red, draw=red, scale = 2}, forget plot]
  table[row sep=crcr]{%
12	57.6657820508611\\
};
\addplot [color=red, only marks, mark=*, mark options={solid, fill=red, draw=red, scale = 2}, forget plot]
  table[row sep=crcr]{%
13	53.9405244332119\\
};
\addplot [color=red, only marks, mark=*, mark options={solid, fill=red, draw=red, scale = 2}, forget plot]
  table[row sep=crcr]{%
14	55.2932301607831\\
};
\addplot [color=red, only marks, mark=*, mark options={solid, fill=red, draw=red, scale = 2}, forget plot]
  table[row sep=crcr]{%
15	50.9310139464078\\
};
\addplot [color=red, only marks, mark=*, mark options={solid, fill=red, draw=red, scale = 2}, forget plot]
  table[row sep=crcr]{%
16	48.4747240585293\\
};
\addplot [color=red, only marks, mark=*, mark options={solid, fill=red, draw=red, scale = 2}, forget plot]
  table[row sep=crcr]{%
17	56.3651545576538\\
};
\end{axis}

\end{tikzpicture}%
}
         \caption{Oxflower17}
         \vspace{0.1cm}
     \end{subfigure}
     \\ 
     \begin{subfigure}[b]{0.8\columnwidth}
         \centering
         \resizebox{\textwidth}{!}{
         \input{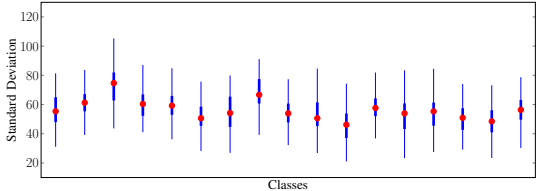}}
         \caption{Oxford-IIIT-Pet}
         \vspace{0.1cm}
     \end{subfigure}
     \\ 
     \begin{subfigure}[b]{0.8\columnwidth}
         \centering
         \resizebox{\textwidth}{!}{
         \input{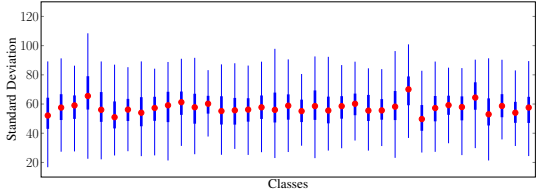}
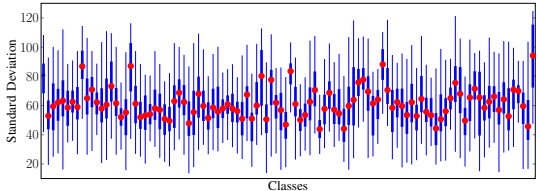}
         \caption{Caltech101}
     \end{subfigure}
	\caption{Box plot of the standard deviation for every image of the three examined datasets. The data points are grouped according to the respective image class. The red marks indicate the median value of the standard deviation of all images in the same class. The thin blue lines indicate the extend of the standard deviation within each class reaching from the minimum to the maximum value. The lower and upper end of the thick blue line indicate the 25th and 75th percentile, respectively.}
	\label{fig:variance}
\end{figure}

 
The evaluation results for the \textit{Caltech101} dataset can be seen in Table~\ref{tab:caltech}. The structure of the table is again the same as for the two previous datasets. Again the best test result for each trained network model is given in bold. The numbers given in Table~\ref{tab:caltech} suggest to use bilinear interpolation during the pre-processing of the dataset whenever a non frequency based interpolation method is used during training. \textit{Caltech101} dataset has a large number of available training samples and a high image variance both between the different classes as well as within the same class. This can be seen from the box plots in Figure~\ref{fig:variance} where the standard deviation of each image per class is plotted for all three examined datasets. While for the \textit{Oxflower17} and \textit{Oxford-IIIT-Pet} dataset the median values of the image standard deviation, indicated by the red marks in Figure~\ref{fig:variance}, are close together for the single classes of the dataset, the variation of the median values among the different classes is much higher for the \textit{Caltech101} dataset. Figure~\ref{fig:variance} also shows that the variance of the standard deviation of the images within the single classes is much higher for the \textit{Caltech101} dataset than for the two other datasets. Moreover, the value range of the standard deviation within the same class is more widespread for \textit{Caltech101} than for \textit{Oxflower17} and \textit{Oxford-IIIT-Pet} which is marked by the thin blue lines in Figure~\ref{fig:variance}. Therefore, the training process benefits more from a smoother interpolation assimilating the single images as done by bilinear interpolation, than from remaining small details of the image content as good as possible. Thus, the results for bilinear pre-processing rank above the results with the high quality FSMR and bicubic interpolation. Only the networks trained on FSMR pre-processed data benefit from a test interpolation method with higher visual quality i.e. bicubic interpolation or best FSMR. \par %
Our experiments reveal that the usage of FSMR for the interpolation tasks on the test dataset yields better results independent of the preceding interpolation used on the training dataset for small datasets and datasets with small inter-class variance. Here the restoration of small details is very important for the overall classification result. The importance of remaining every small detail decreases for networks with a small receptive field trained on large datasets, as well as for datasets with a large inter-class variance. This is a valuable insight as for most 'out-of-the-box' neural network applications, where pretrained models are used to solve the users problem, the initial interpolation used on the training dataset is unknown. Our experiments suggest the usage of FSMR in cases with a small inter-class variance for the interpolation tasks during the application.
\section{Conclusion}
\label{sec:conclusion}
We propose to use high quality FSMR for neural network applications. In contrast to common interpolation methods as bilinear or bicubic interpolation, FSMR builds a model based on the superposition of weighted basis functions. This interpolation method is capable of maintaining small details within the images and outperforms common interpolation methods in terms of PSNR and SSIM. In our work, we demonstrate that this is especially helpful for small datasets with a small inter-class variance where the correct classification depends on remaining small details. 
We examined four state-of-the-art neural networks from a wide range of number of parameters reaching from 5.3 million parameters for EfficientNetB0 to 60 million parameters for ResNet152. We are able to show that the highest classification accuracy is obtained on test sets processed using the FSMR method for datasets with a small inter-class variance. As the interpolation method used in the initial training is often not known, FSMR offers a technique that may achieve best classification results independent from the preparation of the initial training setup.
\section*{Acknowledgment}
The authors gratefully acknowledge support by the German Federal Ministry of Education and Research (BMBF) under Grant No. 13N15319 and by the Deutsche Forschungsgemeinschaft (DFG) under project number 402837983. 

\pagebreak
\bibliographystyle{IEEEtran}
\bibliography{Interp4NN}

\end{document}